\documentclass[10pt,english,american,twocolumn]{article}
\usepackage{amsmath}
\usepackage{graphicx}
\DeclareMathOperator*{\argmax}{\arg\!\max}
\begin{document}
\title{Moving Toward High Precision Dynamical Modelling in Hidden Markov Models}%
\author{S\'ebastien~Gagnon, and~Jean~Rouat PhD,\\ 
sebastien.gagnon2@usherbrooke.ca, jean.rouat@usherbrooke.ca %
\thanks{Thanks to S. Brodeur and S. Wood, for fruitful discussions and to J.-F. Duval for his insightful review. Partial Funding by NSERC discovery}
\thanks{J. Rouat and S. Gagnon are with the NECOTIS research
group, dept. of ECE, Univ. de Sherbrooke, Qc., Canada, J1K 2R1}
}
\maketitle
%
%%%%%%%%%%%%%%%%%%%%%%%%%%%%%%%%%%%%%%%%%%%%%%%%%%%%%%
%		Abstract
%%%%%%%%%%%%%%%%%%%%%%%%%%%%%%%%%%%%%%%%%%%%%%%%%%%%%%
\begin{abstract}
Hidden Markov Model (HMM) is often regarded as the dynamical model of choice in many fields and applications. It is also at the heart of most state-of-the-art speech recognition systems since the 70's. However, from Gaussian mixture models HMMs (GMM-HMM) to deep neural network HMMs (DNN-HMM), the underlying Markovian chain of state-of-the-art models did not changed much. The ``left-to-right" topology is mostly always employed because very few other alternatives exist. In this paper, we propose that finely-tuned HMM topologies are essential for precise temporal modelling and that this approach should be investigated in state-of-the-art HMM system. As such, we propose a proof-of-concept framework for learning efficient topologies by pruning down complex generic models. Speech recognition experiments that were conducted indicate that complex time dependencies can be better learned by this approach than with classical ``left-to-right" models.
\end{abstract}
%
%%%%%%%%%%%%%%%%%%%%%%%%%%%%%%%%%%%%%%%%%%%%%%%%%%%%%%
%		Introduction
%%%%%%%%%%%%%%%%%%%%%%%%%%%%%%%%%%%%%%%%%%%%%%%%%%%%%%
\section{Introduction}
The correctness of a hidden Markov model's (HMM) topology can strongly influence the model accuracy of a HMM systems, especially for signals with high dynamic variability. This graphical architecture is usually
hand-designed to a simple and generic form (usually shared across all classes), whereas constructing precisely tuned class representations can be
challenging.\par
In the 70's a ``left-to-right" topology was first proposed for speech modelling, meaning that feature changes through time always flowed in a specific sequential order \cite{jelinek}.
It is however simplifying considering that spontaneous speech dynamics are known to be very variable \cite{greenberg}.
Up to these days, most state-of-the-art ASR systems such as \textit{deep neural networks}-HMMs (DNN-HMMs, \cite{mohamed})
are still based on that architecture.\par
As states of HMMs encode static feature space distributions, simple HMM topologies can only model coarse dynamics.
A dynamic process constructed from static events is as detailed as the number of such events. Too little precision results in an
\textit{underfitted} model with low discriminative power in classification systems. Too much precision, however, can lead
to an \textit{overfitted} model without generalizing power \cite{cawley}, making it unable to recognize anything but the training signals.
In model selection, as discussed in \cite{cawley}, the key is to balance precision and generalization for maximum performance.\par
\textit{Robustness}, the ability of a system to tolerate recording environment changes, is also to be considered and seems to be strongly
related to a model's generalizing power \cite{xiong}. As such, improving model precision would decrease robustness. \textit{Online adaptation} techniques, however, can easily compensate for such
a drawback. Based on a more comprehensive approach, these methods deal with unseen noise by modifying model statistics at testing time \cite{ozlem,li,narayanan}.
Performance of such systems in noisy conditions are quite remarkable.\par
Even if ``left-to-right" topologies are good for speech signals, some datasets with higher temporal complexity need more precise architectures.
The fact, however, is that topologies are usually hand-designed and kept simple, following Occam's razor principle \cite{mackay}. Therefore, the main goal of
this work is to provide a framework to automatically learn precise HMM topologies from data.\par
``Left-to-right" HDP-HMM \cite{torbati}, a recently developed technique, is one example of such a framework; it is however unbound by Occam's razor principle or any
strong underfitting/overfitting criterion. The expected size of the learned topology being dependant on a \textit{concentration} parameter chosen \textit{a priori},
the designer has indirect control over the degree of temporal precision. Furthermore, the model is allowed to both increase or decrease its size according
to a Dirichlet process and is thus unconstrained to follow Occam's principle \cite{mackay}. On the other hand, this pioneer work allow for the development of HMM speech models without
any \textit{a priori} knowledge of the dynamics, something that is not possible with other approaches.\par
While adept at learning complex topologies from data, ``left-to-right" HDP-HMM is oblivious to recognition accuracy and how the architecture influences it.
Like most HMM training procedures, this is caused by a mismatch between training and decoding alignement methods, i.e. \textit{forward-backward} vs \textit{Viterbi}.
In \cite{torbati}, for example, some learned monophone topologies allow decoded paths to be as short as 2 time frames long whereas in standard monophone systems the shortest path is 3 time frames.
According to our preliminary experiments, this tend to generate insertion errors when decoding, enough to significantly lower accuracy as defined by the word error rate (WER) standard.
As shown in TIMIT benchmarks listed in \cite{lopes}, considering insertion errors always significantly decreases the recognition performance of a system. Thus, while
useful in approximating the topology needed for each class, ``left-to-right" HDP-HMM procedure alone is not ideal for our intended goal.\par
Conventional approaches to dynamics encoding with HMMs usually substitute topology learning with transition
probabilities estimation. In speech recognition, this is the most popular paradigm: generic ``left-to-right" architectures are
adapted to target signal's dynamics by tuning a few persistent parameters. In \cite{yong}, such an approach is attempted on a complex
generic topology with improved additive noise robustness. However, clean speech performance are not reported, which may suggests that
precision has not improved. In fact, improved robustness can be linked to a greater generalizing power, as explored in \cite{xiong}.\par
These results might be explained by an intrinsic problem of HMMs, the imbalance between the dynamic ranges of the
transition and emission probabilities. Exposed by Rabiner and Huang in \cite{rabiner}, this phenomenon is at the root of the popular thought that
transition probabilities are almost useless. It is even a common practice for designers to implement HMM ASR systems with
untrained transitions, because the loss in performance is fairly small. Explained in \cite{rabiner} as a lack of pervasive discriminative
power of the transition probabilities in path decoding, we conceptualize its effects as rendering equiprobable all transitions
that leave the same state. Thus, tuning transition probabilities cannot be a good substitution to complex topology learning.\par
In this work, we first analyse the effects of the imbalance phenomenon. We show that all
paths leaving the same state are effectively equiprobable in the standard TIMIT monophone recognition experiment. Thus, topology learning is
shown essential for precise dynamics modelling, for which we then propose a simple and accessible framework. Assuming that HMM spoken word models
in conventional ASR systems are closer to underfitting than overfitting (a reasoning we based on \cite{greenberg}), we propose to use model \textit{flattening}
\cite{yong} in conjunction with \textit{transitions pruning} to extract precise class topologies. \textit{Flattening} is the process of transforming a simple
``left-to-right" \textit{Gaussian mixture model}-HMM (GMM-HMM) into an equivalent complex HMM with single Gaussian emission models. Using \textit{transitions pruning}
to reduce the flatten model complexity then reveals a more precise dynamic model while still following Occam's razor principle.
We finally demonstrate that with the same number of emission model parameters, our technique clearly outperforms the classic ``left-to-right"
topology on clean word recognition tasks.\par
%
%%%%%%%%%%%%%%%%%%%%%%%%%%%%%%%%%%%%%%%%%%%%%%%%%%%%%%
%	Transition and Emission Probabilities Imbalance
%%%%%%%%%%%%%%%%%%%%%%%%%%%%%%%%%%%%%%%%%%%%%%%%%%%%%%
\section{Transition and Emission Probabilities Imbalance}
As discussed in \cite{rabiner}, transition probabilities may not play a significant role in
path decoding (using Viterbi); recognition could be entirely independent of them. Then, all transitions
leaving the same state could be considered effectively equiprobable during path decoding.
To the knowledge of the authors, this phenomenon has not been quantitatively documented for speech recognition. In the
\textit{token passing} implementation of the Viterbi algorithm, the state $s(t+1)$ occupied at time $t+1$ is given by \cite{young}:%
\begin{equation}
s(t+1)=\argmax_{k=1,2,...,N} [a_{s(t) \rightarrow k} b_k (O_{t+1})]
\end{equation}
Where $N$ is the total number of emitting states in the model. Let there be a distinction between zero and non-zero transition
probabilities:%
\begin{equation}
\forall (i,k) \in [1,...,N]; a_{i \rightarrow k} \neq 0 \rightarrow k \in \varphi_{i}
\end{equation}
Were $\varphi_{i}$ regroups all non-zero transitions leaving state $i$. Thus, (1) can be reformulated in the following fashion:%
\begin{equation}
s(t+1) = \argmax_{k \in\varphi_{s(t)}} [a_{s(t) \rightarrow k} b_k (O_{t+1})]
\end{equation}
Equation (3) only takes into account states that are linked
by a non-zero transition from $s(t)$. Formula (3) can be formulated as follows:%
\begin{multline}
s(t+1)=i \; \; if \\ 
\frac{b_i (O_{t+1})}{b_k (O_{t+1})} > \frac{a_{s(t) \rightarrow k}}{a_{s(t) \rightarrow i}},
\forall (i,k) \in \varphi_{s(t)},k \neq i
\end{multline}
\begin{equation*}
\mbox{Lets define:} \;
\beta = \frac{b_i (O_{t+1})}{b_k (O_{t+1})}
\; \;
\alpha = \frac{a_{s(t) \rightarrow k}}{a_{s(t) \rightarrow i}}
\end{equation*}
We defined $\alpha$ and $\beta$ as ratios of probabilities to isolate the respective transition and emission discriminability forces.
The variance of these variables, respectively the transition and emission discriminability coefficients, give a good estimate of their dynamic ranges.
To evaluate them, we conducted an experiment on the TIMIT training set with conventional 5-states (3 emitting states) ``left-to-right"
monophonic models with Gaussian mixture models (GMMs) of 16 components on each state. The procedure is done in 2 steps for each training utterance: first, using the appropriate
models (listed in the signal's label) an ideal path is computed with the \textit{forward-backward} algorithm \cite{rabiner89}. Then, for each transition
taken in the decoded path the $\alpha$ and $\beta$ values are computed. The variances are then estimated across all the training set. The path alignment
method used here, forward-backward, is not equivalent to (4). In fact, Token passing does not take the backward probability into account and is therefore less optimal.
This was done purposefully to favor high emission probabilities in an effort to minimize discriminative power. One must understand that strong model mismatch comes
from emission probability values being several orders of magnitude different from one state to another, which far more happens in low probabilities (in mismatched dynamics). In other words, the emission discriminability
coefficient calculated is minimized to a level unattainable in practical applications.
\begin{equation*}
\sigma(ln(\alpha))=0.80 \; \; \sigma(ln(\beta))=193.24
\end{equation*}
Where $\sigma(i)$ is the variance of $i$. In the linear domain, the standard deviation of $\beta$ is roughly 440,000 times larger than $\alpha$.
Thus, the transition probabilities are in some sense binary variables, i.e. they are, or not, members of $\varphi_{s(t)}$ in (4).\par
This is because emission probabilities have a near-infinite dynamic range, while transition probability do not, for any given $s(t)$. In fact, considering
how this problem is exposed in \cite{rabiner}, we infer that only in topologies with states of near-infinite branching ratios may this imbalance
vanishes. It is therefore safe to assume that all discrete topologies considered in this work are equally affected by this imbalance.%
%
%%%%%%%%%%%%%%%%%%%%%%%%%%%%%%%%%%%%%%%%%%%%%%%%%%%%%%
%		Pruning
%%%%%%%%%%%%%%%%%%%%%%%%%%%%%%%%%%%%%%%%%%%%%%%%%%%%%%
\section{Pruning}
Encoding acoustic dynamic properties in a generic HMM model is to change its topology, i.e. by activating or
deactivating transitions. A deactivated transition has a probability value of 0 and is therefore not involved in (4).\par
Learning the topology can be done in 3 ways: either ``growing" from a simple prototype model (ex. \cite{jitsuhiro}), ``pruning" from a
complex generic model (ex. \cite{mak}) or a mixture of both (ex. \cite{torbati}). With ``growing" techniques, i.e. increasing
the model's complexity, an almost groundless guess must be made to determine how the expansion is done. This is very much subject to human error.\par
On the other hand, ``pruning" processes are much more reliable as one removes only the paths that are not often visited.
Mak and Chan \cite{mak}, for example, have successfully used pruning on a ``left-to-right" topology with long range transitions.
When compared with an unpruned system, they obtained a significant improvement on the accuracy in a clean word recognition task.
Our work follows that line of thinking and implements pruning instead of other alternatives.%
%
%%%%%%%%%%%%%%%%%%%%%%%%%%%%%%%%%%%%%%%%%%%%%%%%%%%%%%
%		Proposed System
%%%%%%%%%%%%%%%%%%%%%%%%%%%%%%%%%%%%%%%%%%%%%%%%%%%%%%
\section{Proposed System}
\subsection{Integration of the Pruning Module and Threshold Optimization}
A modification of the standard HMM training procedure is proposed for increasing the temporal modelling precision. Fig. 1
illustrates the full proposed system. The pruning (step \#5) is done by comparing each individual transition
probability with a threshold value, $\epsilon$:%
\begin{equation}
\mbox{if} \; a_{i \rightarrow j} > \epsilon \; \mbox{then keep}, \;
\mbox{else} \; a_{i \rightarrow j} = 0
\end{equation}
Since the value of $\epsilon$ is unknown, optimization steps are required during training to find $\epsilon$
(using the loop step \#6 in Fig. 1). The very simple $\epsilon$ optimization process is designed as such to exploit the
steady relation between pruning threshold and performance, thus avoiding unpredictable local minimums.\par
\begin{figure}
\centering
\includegraphics[width=0.45\textwidth]{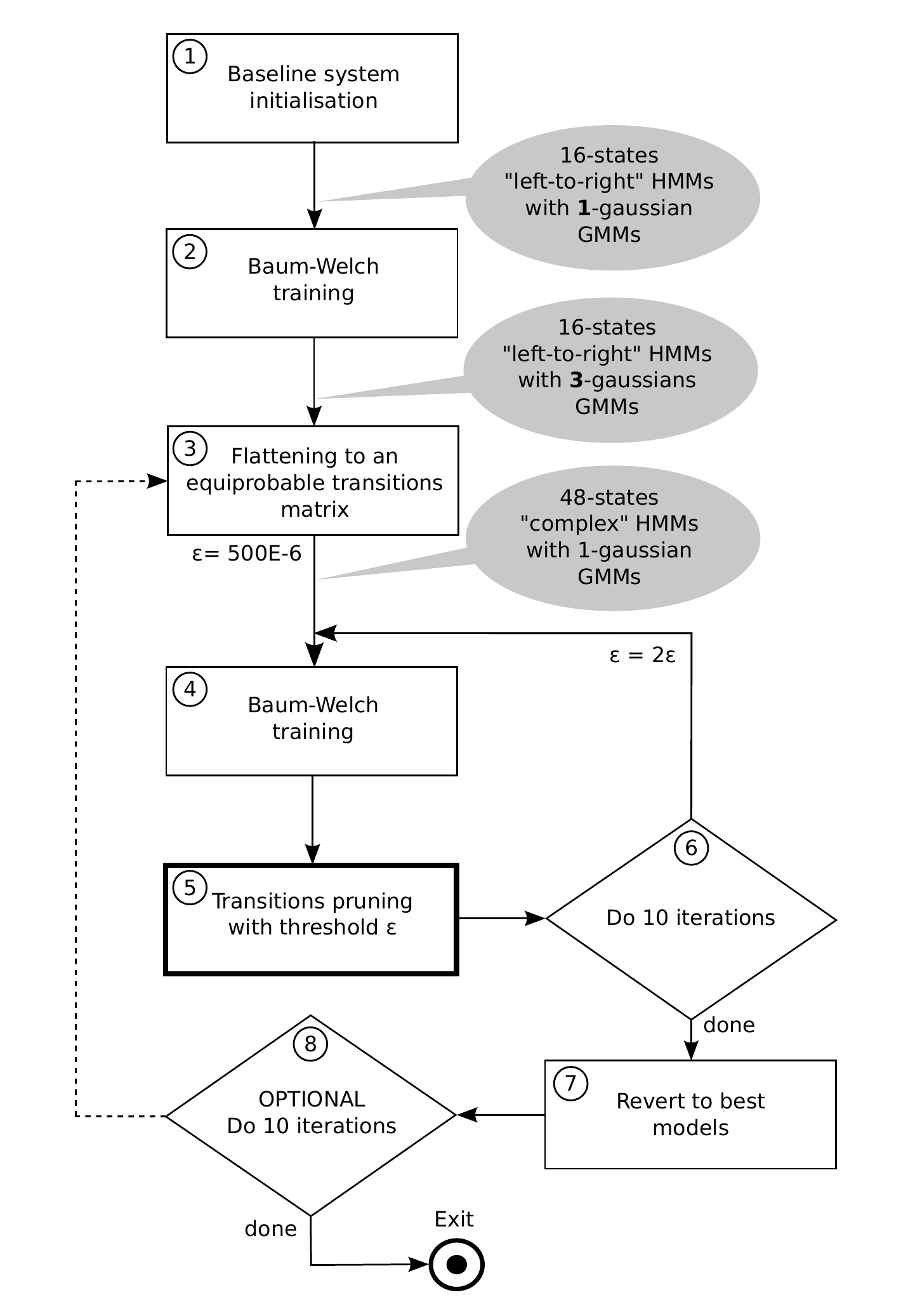}
\caption{1) Flat initialisation; 2) Conventional Baum-Welch training with GMM mixture splitting; 3) Flattening process on the baseline GMM-HMM, transition matrix is equiprobable on the allowed transitions; 4) Baum-Welch training without GMM mixture splitting; 5) Pruning as per (5); 7) The model with highest decoding accuracy on
the training set is kept, all the others are discarded; 8) (Optional) Emission models feedback: emission models of the model kept in 7 are given to their respective states in the flatten model in 3. All pruned
transitions are thus reactivated.}
\end{figure}%
\subsection{Initialization by Model Flattening}
To maximize the beneficial impact of transition pruning we work on a complex prototype model (high number of states and transitions). However, this can be difficult since 
the transition parameter space is larger than with simple ``left-to-right" topologies. Furthermore, if it is improper, alternate paths
tend to die off during training (i.e. very low occupancy probability) to only favour a single path through the topology. This effectively
returns the model to a long ``left-to-right" chain with mediocre performance. Thus, the initialization of a complex
HMM model is an important aspect of this work, for which the \textit{flattening} technique presented in \cite{yong} is used.
Since a multi-gaussian mixture model can be viewed as an HMM of single-gaussian states, the
flattening consists in replacing the states of a ``left-to-right" GMM-HMM by their respective HMM's form. This effectively flattens
the representation to a lattice of mono-gaussian densities, as illustrated in Fig. 2. The final form of a trained ``left-to-right" model
is already fine-tuned to the acoustic properties of the
modelled class, it is therefore an ideal configuration.%
\begin{figure}
\centering
\includegraphics[width=0.45\textwidth]{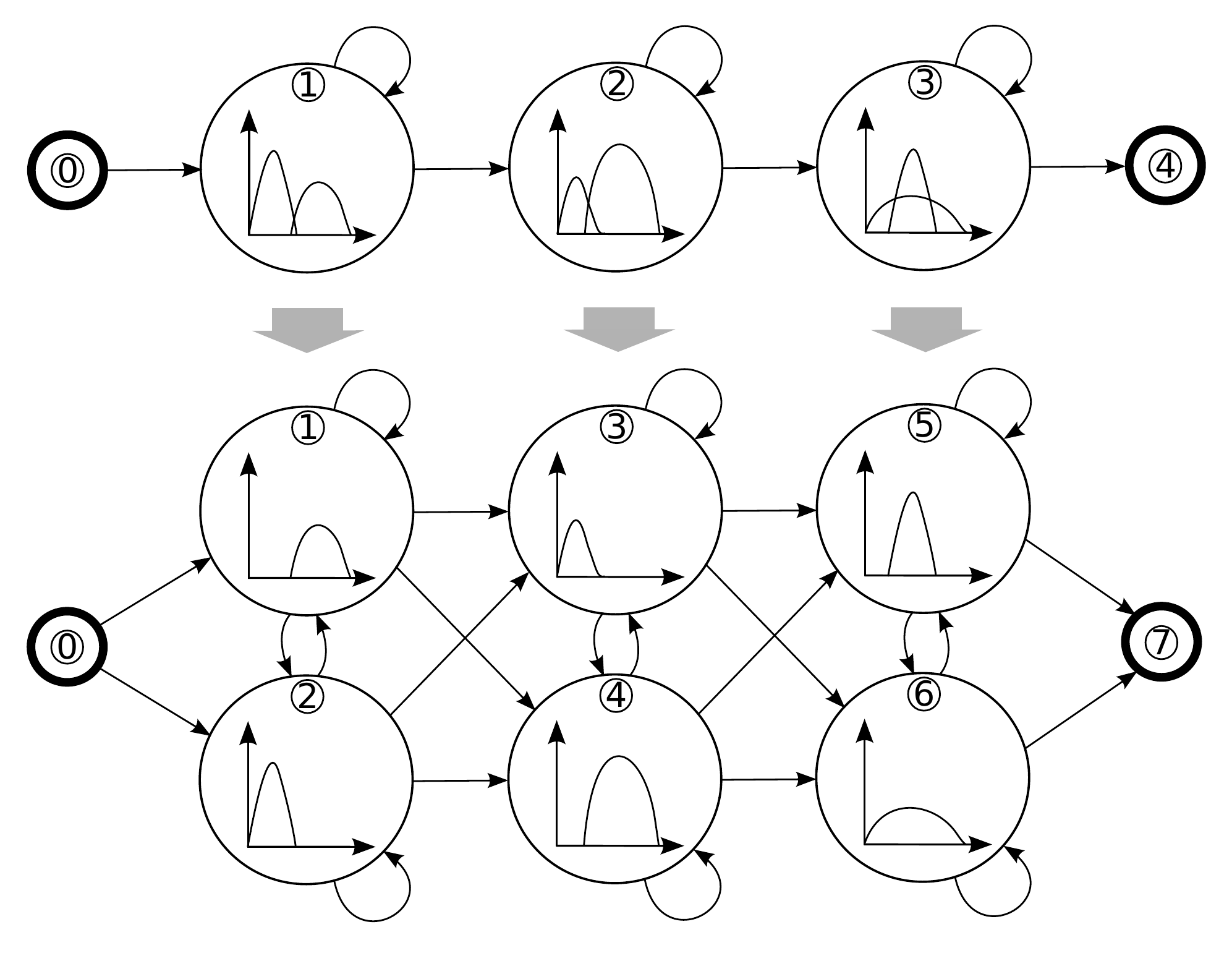}
\caption{Flattening process; the top ``left-to-right" model has a 2 Gaussian mixture model for each state; the bottom ``complex" model is a flatten version of the top model with only 1 Gaussian per state.}
\end{figure}
\subsection{Feedback of Emission Models}
To further increase the recognition performance of the proposed system, the emission models of the pruned
HMMs are fed back to the initialization step (link between steps \#8 and \#3 in Fig.1):%
\begin{equation}
\forall i \in [1,...,N]; b_i^{pre}(\cdot) \leftarrow b_i^{post}(\cdot)
\end{equation}
Where $b_i^{pre}(\cdot)$ and $b_i^{post}(\cdot)$ are respectively the emission distributions of the state $i$ at initialization and after training.
As complex models require more care to train adequately
because of their vast transition parameter space, this step is added to allow slower convergence. We observed experimentally that with about 10 iterations of this ``feedback"
process (step \#8 in Fig. 1) the recognition performance on the training set seems to saturate. While this step is not required
to beat baseline accuracies on clean word recognition, on average it increases performance even further (Table 1). However, it also sharply increases
the computational effort required for training.%
%
%%%%%%%%%%%%%%%%%%%%%%%%%%%%%%%%%%%%%%%%%%%%%%%%%%%%%%
%		Experimental Framework
%%%%%%%%%%%%%%%%%%%%%%%%%%%%%%%%%%%%%%%%%%%%%%%%%%%%%%
\section{Experimental Framework}
To evaluate the temporal modelling precision, we chose to view it in terms of
recognition accuracy. Our premise is that classical ``left-to-right" modelling of speech dynamics is closer to
\textit{underfitting} than \textit{overfitting} and therefore improved precision should yield better results. According to \cite{greenberg},
this is most likely true for spoken word models. Furthermore, clean speech accuracies (the tested signal was recorded in the same
conditions as the training signals) are the main focus since, as explored in \cite{xiong}, noisy recognition seems to deal more with
generalizing power than precision.\par
For all conducted tests, we want any increase in recognition accuracy to be entirely attributed to the
higher temporal modelling precision. This is done by ensuring that the total amount of Gaussian distributions
for each HMM was the same, i.e. the summed amount of GMM mixture components across each model is identical for
both the reference and the proposed systems.\par
First, large dictionary speech recognition applicability is evaluated with a word recognition task with monophonic models on
TIMIT. The baseline ``left-to-right" 5-states (3 emitting states) models have 10 Gaussian mixture components per
state. The proposed system uses flatten versions of the ``left-to-right" models and are thus 30 states long with
single-Gaussian GMMs.\par
Next we tested on the Aurora-2 digits word classification task. Since monophonic
models should not posses much temporal structure, by linguistic definition, a word recognition task is thought to be more representative
of the difference in temporal modelling precision (word models encode much richer dynamics than monophones). The baseline ``left-to-right" HMM models used for the Aurora-2
tasks are 18-states long (16 emitting states) with 3-Gaussian GMMs. The proposed models are 50-states long
(48 emitting states) with 1-Gaussian GMM per state. Training is done on the Aurora-2 clean speech ``TRAIN" corpus. Noisy
recognition tasks are also performed to evaluate the robustness of the technique.\par
Since the clean speech recognition accuracies of the reference ``left-to-right" GMM-HMMs on this last dataset
are very high ($>$99\%), improvements may not be significant. As a result, we generated 4 new versions of the Aurora-2
digits dataset, each of them convoluted with a different real world reverberation impulse response (IR) taken from
the Openair IR database \cite{openairlib}. The IRs
are chosen on the basis of their uniqueness. We selected ``Maes Howe", ``Falkland Palace Royal Tennis Court",
``Purnode's Tunnel" and ``Tyndall Bruce Monument". For these experiments, both systems are trained on a convoluted
version of the ``TRAIN" corpus. Clean and noisy testing datasets are also convoluted in the same fashion, which
means the latter is corrupted by IRs with its additive noise.\par
Word error rates (WER) are computed in the standard fashion, taking into account insertion and deletion errors.
For every one of the 3 Aurora-2 sets (A, B, and C), presented WER values are averaged over all additive noise types. Furthermore,
noisy tests are averaged over SNRs 20, 15, 10, 5 and 0 dB.
%
%%%%%%%%%%%%%%%%%%%%%%%%%%%%%%%%%%%%%%%%%%%%%%%%%%%%%%
%		Results and Discussion
%%%%%%%%%%%%%%%%%%%%%%%%%%%%%%%%%%%%%%%%%%%%%%%%%%%%%%
\section{Results and Discussion}
\begin{table}
\centering
\caption{Performances measured on the Aurora-2 and TIMIT datasets}
\includegraphics[width=0.48\textwidth]{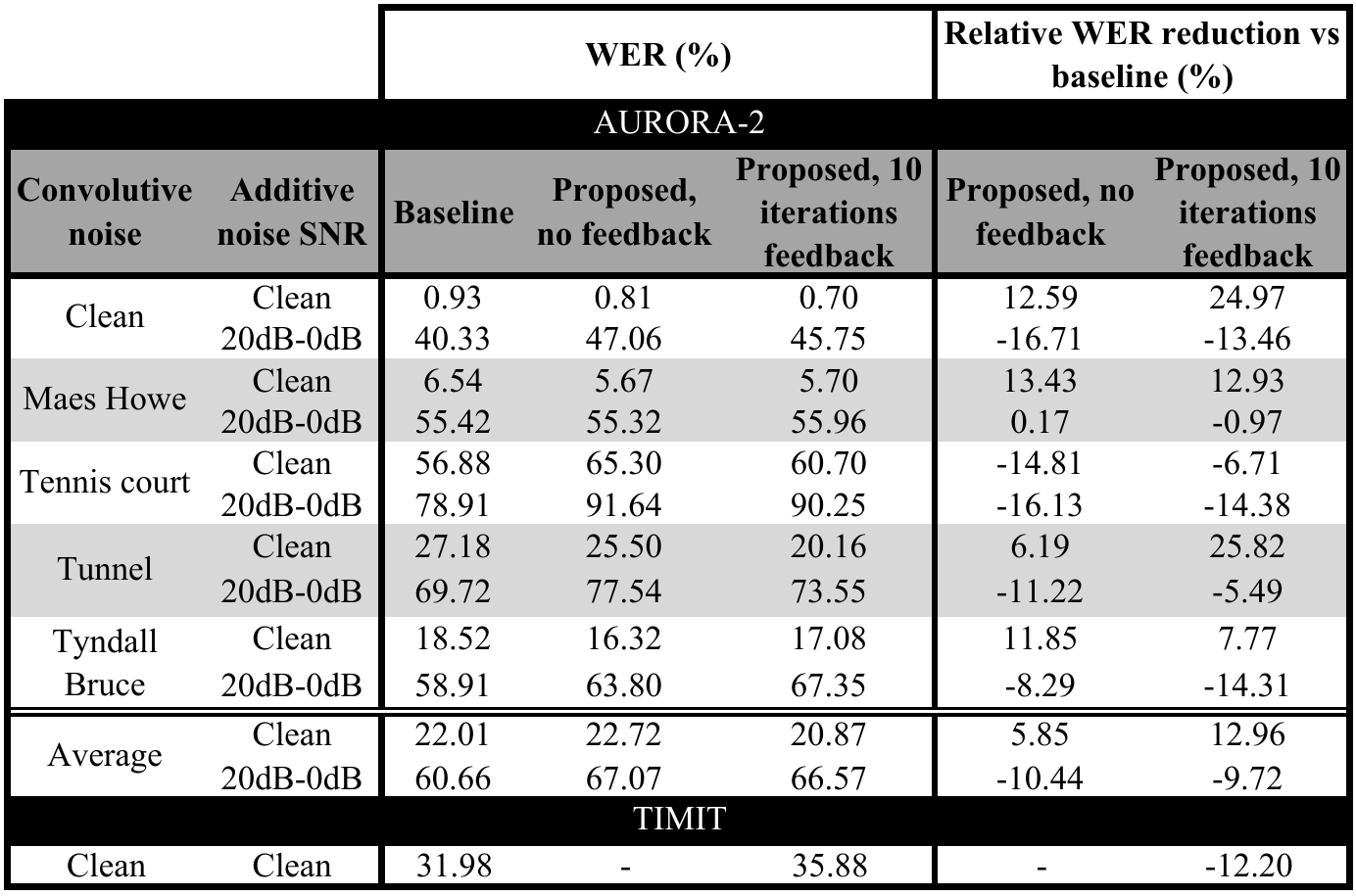}
\end{table}
Table 1 shows significant WER reductions with the proposed approach in clean word recognition. When trained and
tested in reverberated environments without additive noise, performance are also significantly
improved on average. Since reverberation adds time-dependencies to signals and thus increases their
temporal complexity, this demonstrates that temporal precision has indeed been increased. However, the proposed system is
also less robust to additive noise. This may be explained by
a loss of generalizing power caused by the improved precision \cite{xiong}.
To our knowledge, the poor performance with the TIMIT database (shown in Table 2) are best explained by the fact that
monophonic HMMs have linguistically little to no temporal structure, as the name implies. Hence, a model specially
designed to encode complex temporal behaviours is unsuited to this recognition task.\par
We thus see that monophone-based speech recognition does not seem to profit from models with higher temporal
precision while word-based classification does. As discussed in \cite{greenberg}, this indicates that a high
temporal variability exists at the syllable level. Considering this, we suggest that the proposed approach should perform better in large dictionnary systems with classes representing linguistic units of increased length, such as triphones \cite{lopes}.\par
The proposed proof-of-concept framework could be improved in a number of ways for hypothetical increased performances. First, a less trivial optimization algorithm (see Fig.1) could be used in the pruning iterative mechanism. This could yield very precise pruning strengths leading to high recognition accuracy.\par
Better initialization models and methods could also greatly benefit our solution. As it was discussed earlier, complex and finely-tuned models are very sensitive to their initialization conditions and as such,  there is much work to be done in optimizing them. Finally, the proposed framework can also be coupled with complementary state-of-the-art techniques that implement better emission models such as DNN-HMMs \cite{mohamed} and online model adaptation \cite{li}.
%
%%%%%%%%%%%%%%%%%%%%%%%%%%%%%%%%%%%%%%%%%%%%%%%%%%%%%%
%	Acknowledgments and bibliography
%%%%%%%%%%%%%%%%%%%%%%%%%%%%%%%%%%%%%%%%%%%%%%%%%%%%%%
%\section*{Acknowledgements}
%The authors would like to thank Simon Brodeur and Sean Wood, from the NECOTIS research group, for fruitful discussions. We would also
%thank Jean-Fran\c cois Duval for his insightful review. Partial Funding by NSERC discovery.%
%
%\newpage
\bibliography{ref}
\bibliographystyle{IEEEbib}
\end{document}